\documentclass[sigconf]{acmart}

\usepackage{booktabs} % For formal tables
\usepackage{algorithm}
\usepackage{algorithmic}
\usepackage{caption}
\usepackage[lofdepth,lotdepth]{subfig}

\usepackage{fontawesome}
\setcopyright{rightsretained}
\copyrightyear{2018} 
\acmYear{2018} 
\setcopyright{acmcopyright}
\acmConference[GECCO '18]{Genetic and Evolutionary Computation Conference}{July 15--19, 2018}{Kyoto, Japan}
\acmBooktitle{GECCO '18: Genetic and Evolutionary Computation Conference, July 15--19, 2018, Kyoto, Japan}
\acmPrice{15.00}
\acmDOI{10.1145/3205455.3205584}
\acmISBN{978-1-4503-5618-3/18/07}

\begin{document}
\title{On an Immuno-inspired Distributed, Embodied Action-Evolution cum Selection Algorithm
}

\author{Tushar Semwal}
\affiliation{%
  \institution{Indian Institute of Technology Guwahati}
}
\email{t.semwal@iitg.ac.in}

\author{Divya D Kulkarni}
\affiliation{%
  \institution{Indian Institute of Technology Guwahati}
}
\email{divyadk@iitg.ac.in}

\author{Shivashankar B. Nair}
\affiliation{%
  \institution{Indian Institute of Technology Guwahati}
}
\email{sbnair@iitg.ac.in}

\begin{abstract}
Traditional Evolutionary Robotics (ER) employs evolutionary techniques to search for a single monolithic controller which can aid a robot to learn a desired task. These techniques suffer from \textit{bootstrap} and \textit{deception} issues when the tasks are complex for a single controller to learn. Behaviour-decomposition techniques have been used to divide a task into multiple subtasks and evolve separate subcontrollers for each subtask. However, these subcontrollers and the associated subcontroller arbitrator(s) are all evolved off-line. A distributed, fully embodied and evolutionary version of such approaches will greatly aid online learning and help reduce the reality gap.
In this paper, we propose an immunology-inspired embodied action-evolution cum selection algorithm that can cater to distributed ER. This algorithm evolves different subcontrollers for different portions of the search space in a distributed manner just as antibodies are evolved and primed for different antigens in the antigenic space. Experimentation on a collective of real robots embodied with the algorithm showed that  a repertoire of antibody-like subcontrollers was created, evolved and shared \textit{on-the-fly} to cope up with different environmental conditions.
In addition, instead of the conventionally used approach of broadcasting for sharing, we present an \textit{Intelligent Packet Migration} scheme that reduces energy consumption.      
\end{abstract}




\maketitle

%\input{content}

%Tentative title: \textbf{From Generality to Specificity}\\

\section{Introduction}
While in the biological realm, survival may be the only intrinsic motivation behind \textit{evolution}, in Evolutionary Robotics (ER) \cite{nolfi1998evolutionary}, the accomplishment of a set of tasks forms an additional component. These two motives together are deemed necessary in both single and multi-robot scenarios. Due to their inherent resistance to noise and smooth input to output mapping, several researchers have used Artificial Neural Networks (ANN) \cite{floreano3automatic} to evolve controllers for such evolutionary robots \cite{nelson2009fitness}.

Traditional ER approaches incorporate a series of an evolution-guided search for a single \textit{monolithic} robot controller which can aid a robot to achieve the desired tasks. The search for a single optimal robotic controller is performed using several iterations that involve multiple runs of the associated algorithm on centralized (and sequential) simulation environments. The controller is then embedded in real robots. Single controllers have been evolved to implement different tasks such as obstacle avoidance, gait learning and search tasks \cite{nelson2009fitness}. However, bootstrapping the evolutionary process becomes difficult especially when the tasks to be learned are complex for a single controller to cater to. In such scenarios where a complex task may contain bootstrap \cite{gomez1997incremental} and deception \cite{whitley1991fundamental} issues, researchers have opted to use \textit{behaviour-decomposition} \cite{silva2016open} based techniques. In behaviour-decomposition methods, separate subcontrollers are evolved which can cater to different subtasks. An arbitrator that sits on top of these subcontrollers and performs the task of selecting the appropriate subcontroller for the current task \cite{duarte2012hierarchical,lee1999evolving}, is evolved. 

In contrast to traditional ER approaches, the term Embodied Evolution (EE) applies to a collective of robots that autonomously and continuously adapt their behaviour in accordance with the changes in the environment \cite{bredeche2017embodied}. In EE, robot controllers learn both onboard and online, and continue to do so even when the robots are deployed in an environment, thereby reducing the \textit{reality gap} \cite{jakobi1997evolutionary}. EE is still susceptible to bootstrapping and deception issues when the task to be accomplished is complex for single-controller based robots to learn. While under traditional approaches, there are evidence of earlier ER work that use behavioural decomposition \cite{lee1999evolving} to avoid the bootstrap issues but these techniques have been seldom applied in an \textit{embodied} (online and onboard) manner. What needs to be explored is a technique which is a distributed, fully embodied and evolutionary version of such traditional approaches. 

In this paper, we propose an immunology-inspired embodied action-evolution cum selection algorithm to evolve different subcontrollers for different regions of the search space. Sensor values sampled from the environment form the antigen while the associated subcontroller (actions) corresponds to an antibody. Similar to the way antibodies are evolved and primed for different antigens, the algorithm evolves and selects different subcontrollers for an associated region of antigenic (sensory) space. Some of the salient features of the proposed algorithm are:\\
(1) On-the-fly and Onboard Evolution: The subcontrollers are evolved on-the-fly and onboard the robots.\\
(2) Distributed Learning: While the encapsulated version of algorithm within one robot seeks and learns to choose the best subcontroller, sharing of these controllers across peers in the collective of robots speeds up the convergence process \cite{bredeche2017embodied}. \\
(3) Single Parameter Tuning: A single system parameter (explained later) can be used to tune and vary the granularity of search in the antigenic space.

This paper describes an alternative algorithm to evolve and select different subcontrollers on-the-fly for different ranges (portions) of sensory values (antigenic space) sampled from the environment, in a distributed and decentralized manner. The use of an immunology based algorithm ushers a new era in ER wherein multiple subcontrollers can be created, evolved and arbitrated online in a collective of embodied robots.

\section{Related Work}
Though a copious amount of research on ER can be cited, in this section we discuss the behaviour-decomposition and embodied evolution based techniques which are more pertinent to the work presented in this paper.\\

\subsection{Behavioural decomposition}
In this method, instead of just one robot controller, different subcontrollers are evolved to solve distinct subtasks. After these subcontrollers have adequately evolved, another  (usually ANN based) controller is trained to map the input states of the robot to one of the already evolved subtask specific controllers.  Lee \cite{lee1999evolving} describes each of the low-level subtask specific controllers as \textit{behaviour primitives} while the top-level controller that has learned to map the inputs to these subcontrollers was termed the \textit{behaviour arbitrator}. 

Lee \cite{lee1999evolving} implemented a Genetic Programming (GP) based controller to solve a box-pushing task. This task was manually divided into two subtasks. Separate subcontrollers (primitives) were evolved for each subtask on a simulator. Finally, a GP based arbitrator was evolved to combine these subcontrollers hierarchically. Though their approach was implemented on a real robot, each subcontroller was evolved separately in an offline manner. Further, there is no evidence that there proposed architecture can be used across multi-robot scenarios in a decentralized manner. Moioli et al. \cite{moioli2008towards} proposed the GasNet system where subcontrollers for two different tasks were activated or inhibited based on the production and secretion of virtual hormones. Their homeostasis-inspired controller was able to select appropriate subcontrollers depending on internal and external stimuli.

Duarte et al. \cite{duarte2012hierarchical} portrays a search and rescue task using a real e-puck robot. The overall task was divided into a few subtasks and separate subcontrollers were evolved for each of them. The experimenter provided the fitness function for each subtask. After the appropriate subcontrollers were found, a behaviour arbitrator was evolved which delegates a subtask to the best subcontroller based on the sensory inputs. Though a complex task was solved using their hierarchical controller, the behaviour primitives and the behaviour arbitrator do not seem to have been evolved in an online and on-board manner. A similar line of work by Duarte et al. \citep{duarte2014hybrid} also demonstrates an approach for the incremental transfer of their evolved hierarchical control system from simulation to real robots. Of late, Duarte et al. \cite{Duarte:2016:EER:2908812.2908855} have presented EvoRBC, an approach to evolve a control system for robots with arbitrary locomotion complexity and implemented it on a simulated robot. They have used a Quality Diversity algorithm to build a repertoire of behaviour primitives (for e.g. move straight, turn right). This repertoire is then used to evolve an ANN which maps the sensory inputs of the robot to an appropriate primitive. 

In most of the above-reported approaches, the subcontrollers are evolved separately and lack the essence of continuous, distributed, decentralized and on-the-fly learning.

\subsection{Embodied Evolution}
A recent comprehensive paper by Bredeche et al. \cite{bredeche2017embodied} presents a review of the research published since the inception of the term Embodied Evolution (1997-2017). EE involves continuous and online learning in a collective of robots. A population of robots learns in a decentralized manner by sharing the controllers evolved among the peer robots. Although there has been a recent surge in papers \cite{bredeche2017embodied} where EE has been successfully applied, most of the work that cites the use of real robots \cite{watson2002embodied,heinerman2016line,simoes2001embedding,o2014distributed} are constrained to a single controller that can solve relatively simple tasks such as obstacle avoidance and phototaxis. Only recent, a work by Heinerman et al. \cite{heinerman2016line} implements a relatively complex task of foraging using a single controller.     
A robot may require learning several such tasks. Under such conditions, the subcontroller will need to be re-trained to take in the set of new tasks. In an ANN-based subcontroller, such re-training may not be a viable exercise. 
Intuitively, one may conclude that evolving a single subcontroller to cater to an ever-increasing number of tasks, is extremely cumbersome if not impossible. Using behaviour primitives and an associated arbitrator may be a logical step to circumvent this drawback. However, the drawback is that for every incremental addition of a new subtask the behaviour arbitrator has to be trained offline all over again.
This calls for an online and onboard continuous learning mechanism for both the subcontrollers and that arbitrator which will consequently lower the reality gap. The algorithm proposed in this paper distinguishes itself from earlier presented techniques in a way that instead of subtask division, it keeps dividing the whole sensor-sampled search area within the given environment, into separate regions. A subcontroller is then evolved on-the-fly for each such region.

\section{Methodology}
The method proposed in this paper is inspired by the novel action-selection mechanism exhibited by the Biological Immune System (BIS). In this section, we initially introduce the immunological metaphors used, followed by a description of the computational counterparts and the explanation of the proposed algorithm.

\subsection{Immune Metaphors}
\begin{figure}[t]
    \centering
    \includegraphics[scale=.4]{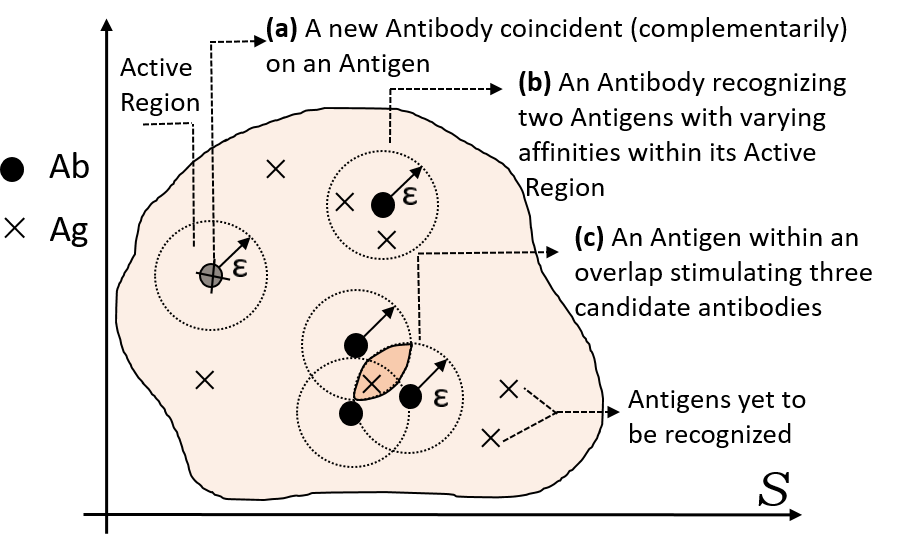}
    \caption{A Shape Space $\mathcal{S}$ } 
    \label{fig:shapeSpace}
\end{figure}

The ability of the BIS to learn and quickly counter the effects of a new or already encountered \textit{antigen} provides tremendous insights and motivation to develop powerful yet simple, selection and learning mechanisms to suit distributed computational scenarios. The BIS comprises several types of immune cells such as B-, T-, Killer and Plasma cells, all of which have \textit{antibodies} to recognize and curb the antigens. Immune cells have monospecific antibodies all around it that perform the main task of antigenic recognition. Though different, for simplicity and brevity, we refer to all the immune cells as antibodies. The term antigen has been used synonymously for all types of foreign pathogens. As soon as an antigen is detected within the body of an individual, the process of generating and/or attracting the right kind of antibodies at the site of detection is initiated. The antibodies make their way to the antigenic site through the fluids present inside the body. \\ 
\textbf{Cross-Reactivity Threshold ($\epsilon$)~: }
Each antibody ($Ab$) has a  \textit{Paratope} ($Pt$) which aids in recognizing an \textit{Epitope} ($Ep$) of the antigen ($Ag$). The extent of the complementarity or the affinity ($\psi$) in the \textit{shapes} of $Ep$ and $Pt$ contributes to a recognition. Greater the value of $\psi$, greater is the potential of that $Ab$ to curb the corresponding $Ag$. An $Ab$ is selected for an $Ag$ only if the complementary shape of the $Ep$ of that $Ag$ lies within a small region of the shape space surrounding that of the $Pt$ of $Ab$. This small region which we refer to as the \textit{Active Region (AR)}, is characterized by the \textit{cross-reactivity threshold} ($\epsilon$) \cite{de2002artificial}. All antibodies within this region are attracted towards the antigen $Ag$ and hence are \textit{stimulated} by it. Stimulated antibodies clone and proliferate proportionate to their affinities \cite{burnet1959clonal} and thus quell the antigenic attack. Fig. \ref{fig:shapeSpace} shows a 2-D shape space $\mathcal{S}$ wherein the black dots signify antibodies and the crosses indicate the antigens. When a newly detected antigen has no antibodies to suppress it, an $Ab$ which is complementarily coincident to the $Ag$ is created (Fig.~\ref{fig:shapeSpace}(a)). All $Ag$s that fall within the active region of this $Ab$ can now be catered to, by this $Ab$. As can be seen in Fig.~\ref{fig:shapeSpace}(b) the $Ab$ can cater to two of the $Ag$s that lie within its $AR$. Several types of $Ab$s could recognize the same $Ag$ if the latter lies within the overlap of their active regions (Fig.~\ref{fig:shapeSpace}(c)). In brief, the role of the BIS is to create, select and evolve a \textit{repertoire} of $Ab$s which are best suited to curb antigenic attacks.  
 
\subsection{From Immunology to the Real World}
\begin{figure}[t]
    \centering
    \includegraphics[scale=.3]{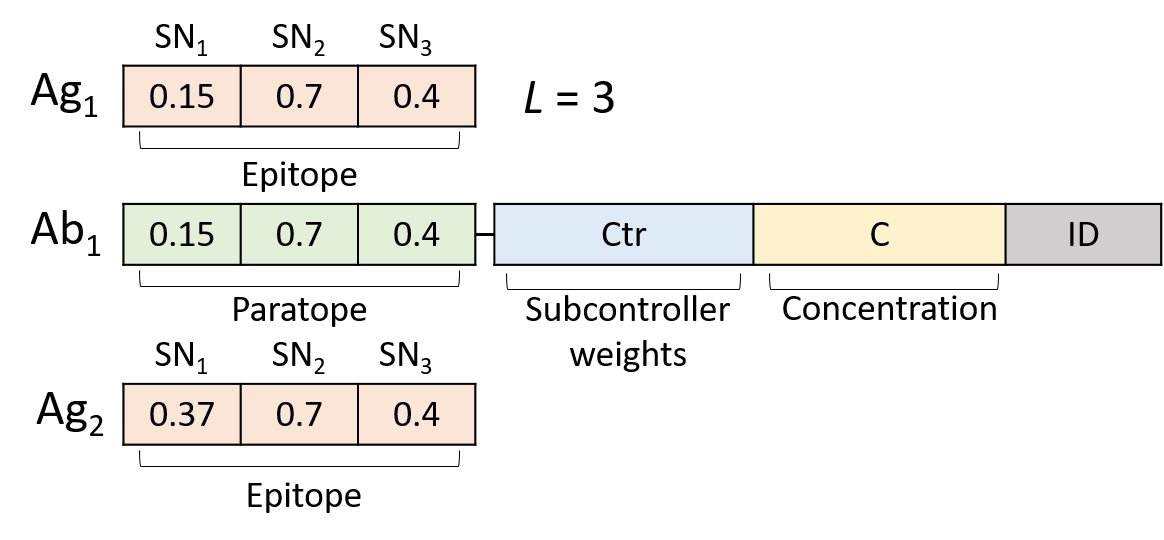}
    \caption{An Antibody tackling two different Antigens ($SN_i$s are the values obtained from sensors onboard the robot)} 
    \label{fig:ag_ab}
\end{figure}

In the real world, the antigenic epitope $Ep$ can be visualized as an $L$-dimensional vector sampled from the environment via sensors on-board a robot. $L$ corresponds to the number of sensors attached to the robot. Fig. \ref{fig:ag_ab} depicts an epitope $Ep$ (top) formed when a robot encounters an obstacle in front. The corresponding antibody $Ab$, as shown in Fig. \ref{fig:ag_ab}, comprises four components - a paratope $Pt$, the associated subcontroller $Ctr$, a concentration $C$ value and an ID number. A new $Ab$ is created if the $Ag$ is not recognized by any of the $Ab$s present in the repertoire. The $Pt$ of such a new $Ab$ is initialized to the $Ep$ of the current $Ag$ so that both $Pt$ and $Ep$ are dimensionally equivalent and have the same shape space. The $Ctr$ is a vector comprising the weights of an associated neural network while the $C$ denotes the fitness value returned after executing the $Ctr$ when the corresponding $Ab$ is chosen to quell $Ag$. The fitness function depends on the task and is provided by the experimenter based on the application. $Ab$s, with better performing $Ctr$s, evolved based on fitness values, are assigned proportionately higher concentrations. The ID number aids in uniquely identifying an $Ab$. Fig. \ref{fig:ag_ab} also shows another $Ag$ (bottom) which falls in the $AR$ of the same $Ab$.
The Euclidean distance between an $Ep$ and a $Pt$ is used as the affinity measure ($\psi$). Lesser this distance more is the affinity. An $Ab$ is chosen to tackle a given environment state ($Ag$) if the affinity $\psi$ between the $Ep$ and the $Pt$ is less than $\epsilon$, a system constant akin to the cross-reactivity threshold \cite{de2002artificial}. Antibodies that satisfy this criterion are referred to as the  \textit{candidate antibodies}. All such candidates antibodies are stimulated by the corresponding antigen $Ag$ located within the overlap of all active regions of these antibodies as shown in Fig. \ref{fig:shapeSpace}(c). This antigenic stimulation ($AgSti$) results in increasing the $C$ of all the candidate antibodies by an amount proportional to their respective $\psi$ values.  In the computational world, a specific subcontroller is synonymous to an $Ab$ while an environmental state acts as an $Ag$. Thus, different subcontrollers could be evolved, each of which is tuned to specific environmental states. In this paper, we present a BIS inspired algorithm to evolve subcontrollers (antibodies) for specific actions required to counter different environment states (antigens) sampled by the sensors onboard a robot. Just as the BIS, this algorithm is distributed in nature and evolves subcontrollers on-the-fly.

\subsection{The Proposed Algorithm}
The proposed algorithm is outlined in Algorithm \ref{algo:1}. The algorithm runs on every robot belonging to the swarm. It works in tandem with a communication routine which facilitates foreign antibodies (subcontrollers)  to be received from the peer robots in the collective and stored in an \textit{extrinsic} antibody repertoire ($XRep$) within the robot. During the task execution by a robot, its \textit{intrinsic} antibody repertoire ($IRep$) is broadcast to all the peer robots within a communication range. Instead of all, only a set $Ab$s from $IRep$ selected based on a fitness criterion can also be  shared.

Initially, the $IRep$ within a robot is a \textit{tabula rasa}. As mentioned, the sensor values sampled from the environment form the epitope $Ep$ of the antigen $Ag$. As soon as $Ep$ of the current $Ag$ has been sampled, the distances ($\psi$) between this $Ep$ and the $Pt$s of all the $Ab$s in  $IRep$ are calculated using the $measureAffinity$ function (line \ref{measureaff} of Algorithm \ref{algo:1}). The $Ab$s for which the $\psi$ values are less than $\epsilon$ are made to be stimulated (using $agStimulation$ function) by the antigen $Ag$ and are then appended to a list of candidate antibodies ($CandAbs$). As in the BIS, the $AgSti$ raises the concentration $C$ of all antibodies within the $CandAbs$ by an amount proportional to their respective $\psi$ values. 
In the beginning, since no $Ab$s exist in the $IRep$, a new $Ab$ is created whose $Pt$ is initialized to $Ep$ and the weights of the associated $Ctr$ are randomly set (line \ref{createnew}). The ID is provided sequentially, starting with 1. The value of its $C$ is set to an initial non-zero minimum value lest it be discarded immediately. Antibodies with $C$ equal to zero, are purged from $IRep$. The new $Ab$ with ID 1 is then added to the currently empty list  $CandAbs$. 
\begin{algorithm}[t]
\footnotesize
\caption{The Proposed Algorithm}
\label{algo:1}
\begin{algorithmic}[1]
\STATE $\epsilon \gets Constant$;
\STATE $IRep,XRep,CandAb \gets \varnothing$;

\WHILE{True}
\STATE $Ep \gets fetchSensorValues()$;
\FOR{each $Ab_i$ in $IRep$}
\STATE $\psi \gets measureAffinity(Ep,Pt_i)$; \label{measureaff}
\IF{$\psi \leq \epsilon$}
\STATE $agStimulation(\psi,Ab_i)$;
\STATE $CandAbs \gets append(Ab_i)$;
\ENDIF
\ENDFOR 

\IF{$CandAb = \varnothing$}
\STATE $NewAb \gets createNewAntibody(Ep)$; \label{createnew}
\STATE $CandAbs \gets append(NewAb)$;
\STATE $IRep \gets append(NewAb)$;
\ENDIF
\IF{$(random() < P_{sharing}$) OR ($XRep.size = 0)$} 
\STATE $BestAb \gets selectBestAntibody(CandAbs)$; \label{selectbest1}
\ELSE 
\STATE $BestAb \gets selectBestAntibody(XRep)$; \label{selectbest2}
\ENDIF
\WHILE{Current $Ag$ space is within the $AR$ of $BestAb$}
\STATE Execute and Evolve $BestAb$ using an EA;
\ENDWHILE
\STATE $CandAb,XRep \gets \varnothing$;
\ENDWHILE
\end{algorithmic}
\end{algorithm}

For final execution by the robot, the best $Ab$ is selected from either $XRep$ or the list of $CandAbs$ based on a probability $P_{sharing}$. The $selecBestAntibody$ function (lines \ref{selectbest1} and  \ref{selectbest2}) selects the matching $Ab$ ($\psi \leq \epsilon$) having the highest $C$ value. The $Ctr$ of the best $Ab$ is evolved in order to produce the adequate behaviour within the associated $AR$. When the robot is executing a controller of the selected $Ab$, the environmental state is continuously sampled. It may be noted that during the task execution, if the $Ep$ sampled from the environment is outside the $AR$ of the currently selected best $Ab$, the current task execution is stopped and the process to select and evolve a new $Ab$ recommences.

The algorithm partitions the environment (antigenic) space sensed by the robot based on $\epsilon$ which in turn defines the area of the \textit{Active Region} ($AR$) of a subcontroller (antibody). More the value of $\epsilon$, more is the environment space catered to by that subcontroller. Increasing $\epsilon$ will mean lesser number of subcontrollers for a given environment space. If $\epsilon$ were to cover the entire environment space then, the algorithm would try to evolve just one subcontroller that can cater to all conditions, as in \cite{heinerman2016line,o2014distributed}. On the other hand, a very low value of $\epsilon$ would mean a smaller $AR$ resulting in too many subcontrollers catering to very specific tasks. \\
\textbf{(1+1)-Online Evolutionary Algorithm: }
The evolution of the $Ctr$ of the best $Ab$ that has been selected can be achieved using an Evolutionary Algorithm (EA). We have leveraged the (1+1)-Online evolutionary strategy \cite{bredeche2009line} to evolve the antibodies with controllers that deliver satisfactory performance. In this strategy, the controller weights are mutated using a Gaussian function with $N(0,\sigma)$, where the value of $\sigma$ doubles if the offspring of the controller of the selected $Ab$ performs lower than the currently used parent controller. An offspring replaces the parent if the former outperforms the latter one.
\begin{figure}[t]	\subfloat[The Robot]{\includegraphics[width = 32mm]{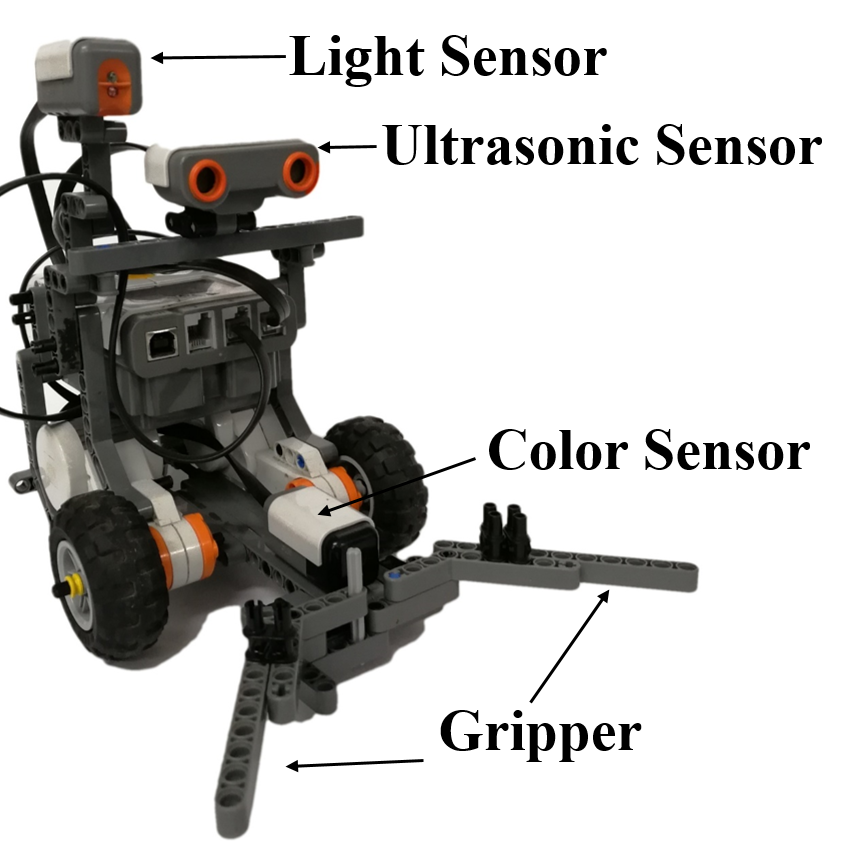} \label{fig:robot}}
\subfloat[The Experimental Arena]{\includegraphics[width = 47mm]{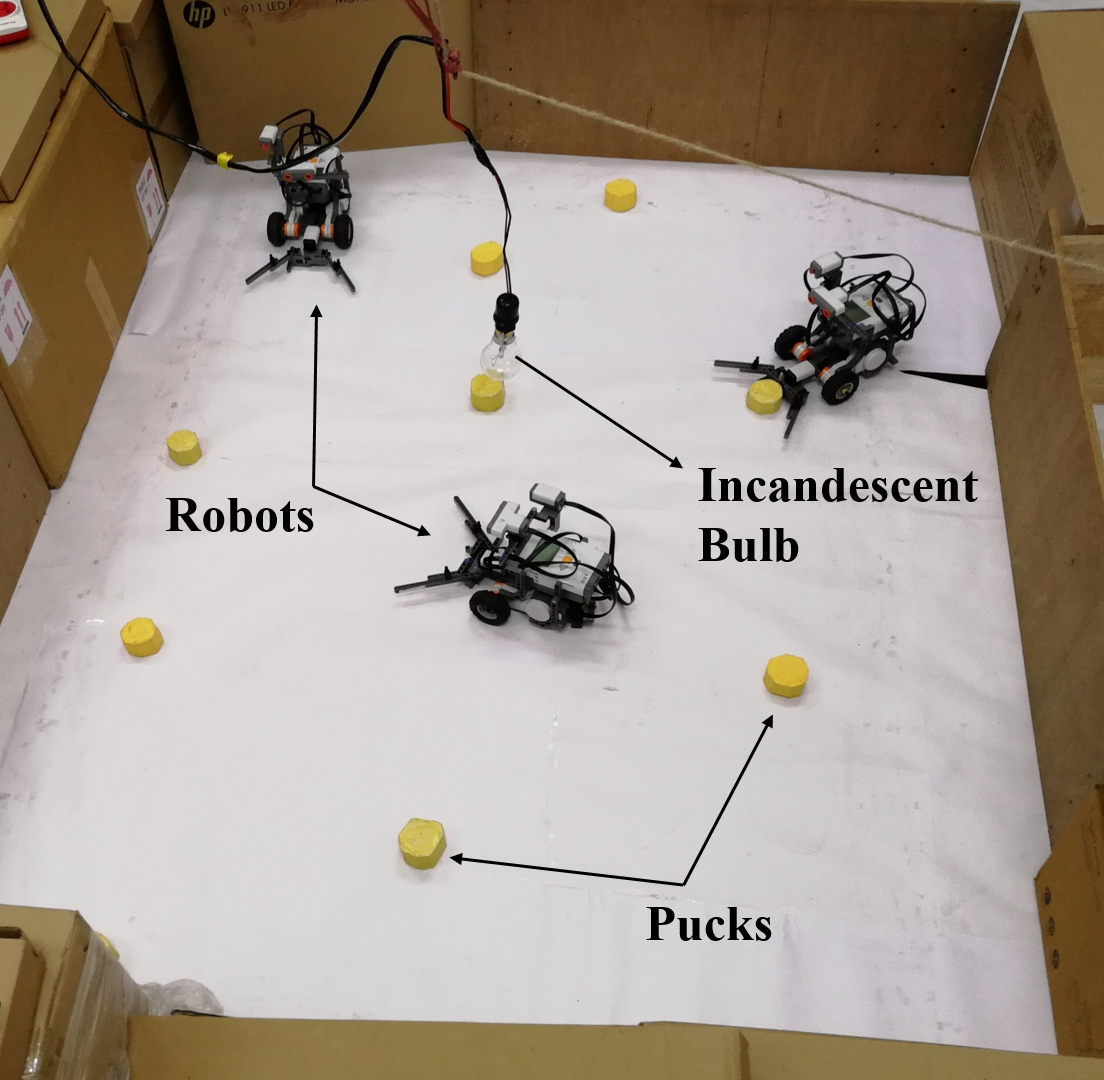} \label{fig:exparena}}
	\caption{(a) Structure of the LEGO\textsuperscript{\textregistered} MINDSTORMS\textsuperscript{\textregistered} NXT robot used in the experiments (b) The Experimental Arena }
	\label{fig:arena}	
\end{figure}

\section{Experiments}
For experimentation, we have used a set of LEGO\textsuperscript{\textregistered} MINDSTORMS\textsuperscript{\textregistered} NXT robots each having a Colour Sensor (CS), an Ultrasonic Sensor (US) and a Light Sensor (LS) as shown in Fig. \ref{fig:robot}. The US mounted in front of the robot aids in obstacle detection while the CS facilitates identifying the colour of an object encountered by the robot. The two LSs are fixed parallel to the two motors in order to detect the light from both directions. The range of values from the ultrasonic, colour and light sensors vary from 0 to 200, 0 to 9 and 0 to 100, respectively. High values from the US denote that the obstacle is far from the robot while low values indicate the presence of the obstacle. High values from the LS show that the robot is near to the light while lower values specify that the robot is far away from the light source. For CS, the default value is calibrated to 7. The high and low values were decided based on the size of the arena. For instance, if the size of the area is more than the detectable range of the US, then the a value near the maximum 200 will correspond to the high values. For tasks involving foraging, a Lego gripper was also attached to the lower frontal part of the robot. With two motors attached to two rear wheels and a caster wheel at the front, the robot used a differential drive to move around in its environment. The operating speed of the motors was set to 70\% of the maximum speed. A Raspberry Pi 3 mounted onboard the robot and interfaced with the sensors and the motors constituted the main hardware controller. As depicted in Fig. \ref{fig:exparena}, 8 to 12 lightweight yellow coloured hexagonal-puck shaped objects were  scattered randomly across a 2m x 2m arena which constituted the environment. For the yellow colour, the CS returned a value 3. The only obstacle in the arena are the walls surrounding the arena. A Wi-Fi router placed near the arena facilitated the interconnection between the robots and hence the sharing of antibodies.  Experiments were performed in both static and dynamic networks. Since Wi-Fi range covers the whole arena, dynamism was implemented by randomly creating and breaking the connection between the robots. The $Ctr$ constituted a feed-forward ANN with four input nodes, five hidden nodes and two output nodes and used the \textit{hyperbolic-tangent} (tanh) as the activation function. Out of the four input nodes, two were connected to the two light sensors, while the remaining were provided inputs from each of the ultrasonic and the colour sensors. The sensor readings were normalized to the range between 0 and 1 and then fed to the inputs of the controller ANN. 

\subsection{Scenarios}
The proposed algorithm was tested on a set of three scenarios explained below in the order of increasing complexity.\\ 
\textit{Scenario $S_1$ - Static Environment}:
In this scenario, a light source (an incandescent bulb) was fixed at some minimum height in the centre of the arena. The objective was to make the robot move towards the light source (phototaxis) and then stay near it. If the robot encountered an obstacle, then it had to avoid it. Using a bare minimum of just a single US and one LS puts additional pressure on the algorithm while searching and evolving an optimum controller. For a single robot controller, the task of phototaxis together with obstacle avoidance thus becomes non-trivial. 
Similar to online evolutionary systems presented in \cite{heinerman2016line}, the objective function used herein rewards behaviours ($Ab$) for each of the sub-goals achieved. The sub-goals herein include motion of the robot towards the light source and avoiding obstacles. The objective function for an evaluation period of $\tau$ timesteps is as follows:
\setlength\abovedisplayskip{0pt}
\setlength\belowdisplayskip{0pt}
\begin{equation}
\label{eq:t1}
\mathcal{F}_{T_1} = \sum_{t=0}^{\tau} (f_{obs} + f_{light})
\end{equation}
\setlength\belowdisplayskip{0pt}
where,$  f_{obs} = v_{trans} * (1 - v_{rot}) * d$ \label{eq:t1a} \\
      $f_{light} = \max\limits_{1 \leq j \leq 2} (lightSensor_{i})$\\ \label{eq:t1b}
$f_{obs}$ is a classical function adapted from \cite{nolfi1998evolutionary} wherein $v_{trans}$ is the translational speed, $v_{rot}$ is the rotational speed and $d$ is the distance between the obstacle and the US on-board the robot. $v_{trans}$, $v_{rot}$ and $d$, are all normalised between 0 and 1. The fitness function  $f_{light}$ rewards movement towards the light source. $lightSensor_i$ is the normalised value from the LS between 0 (no light) and 1 (brightest light).\\
\textit{Scenario $S_2$  - Static Environment with Puck Pushing}: The primary goal here was to push the yellow coloured pucks towards the light source while also avoiding the walls. If no pucks are encountered, the robot should continue moving towards the light source. The CS fixed behind the gripper facilitates the detection of the puck based on its colour. Along with the sub-goals defined in $S_1$, the task $S_2$ also comprises a fitness function to reward the controlled pushing of the puck using the gripper attached to the robot. The objective function for $S_2$ is given by:
\begin{equation}
\label{eq:t2}
\mathcal{F}_{T_2} = \sum_{t=0}^{\tau} (f_{obs} + f_{light} + f_{puck}) 
\end{equation}
where,~ $f_{puck} = b_{puck} + v_{trans}$
\\
$b_{puck}$ is a binary variable which is equal to 1 if a puck is within the gripper else it defaults to 0. The meanings of $f_{obs}$ and $f_{light}$ are same as that defined in scenario $S_1$.\\
\textit{Scenario $S_3$ - Dynamic Environment}: In this scenario, a change in the environment needed to be compensated by a reversal of a behaviour. Here, the light source was switched ON or OFF in an asynchronous manner during run time. When the light was ON, the goal remained the same as in scenario $S_2$ where the robot needed to push the puck towards the light source. However, when the light source was switched OFF, the robot needed to learn to repel these object whenever encountered. Irrespective of whether the light source was ON or OFF, the robot also needed to learn to avoid the walls or any other static obstacle. The objective function used in this scenario is given below.
\begin{equation}
\label{eq:t3}
\mathcal{F}_{T_3} = \sum_{t=0}^{\tau} (f_{obs} + f_{light} + b_{light} * f_{puck} + \overline{b}_{light} * f_{antipuck})
\end{equation}
where, $b_{light}$ is a boolean variable equal to 1 if the light intensity is above a certain threshold. The average of the values returned from the LS at different positions in the arena while the light is ON, forms the threshold. $f_{antipuck}$ models the repulsion of robot when it encounters a puck.

We carried out two sets of experiments - 1) Real-robot 2) Energy Saving. In the real-robot experiment, we switched off the sharing module of our proposed algorithm and used only a single real-robot to learn the respective tasks in the three scenarios explained later. In addition, we also used a set of three real robots to learn the same tasks with the sharing module enabled. This allowed the robots to share their subcontrollers (antibodies) mutually. The second set of experiments were designed with a completely different motive. These experiments were intended to introduce and test a new and energy efficient sharing mechanism. Experiments herein were carried out in an emulated environment with 80 soft or virtual robots connected through a dynamic network. 
Besides, we also performed experiments wherein we attempt to evolve single robot controllers for each of the scenarios. Though not crucial, this was necessary to validate the fact that the used scenarios are substantially complex for a single controller.    

\begin{figure*}[t]    \subfloat[Scenario $S_1$]{\includegraphics[width = 58mm]{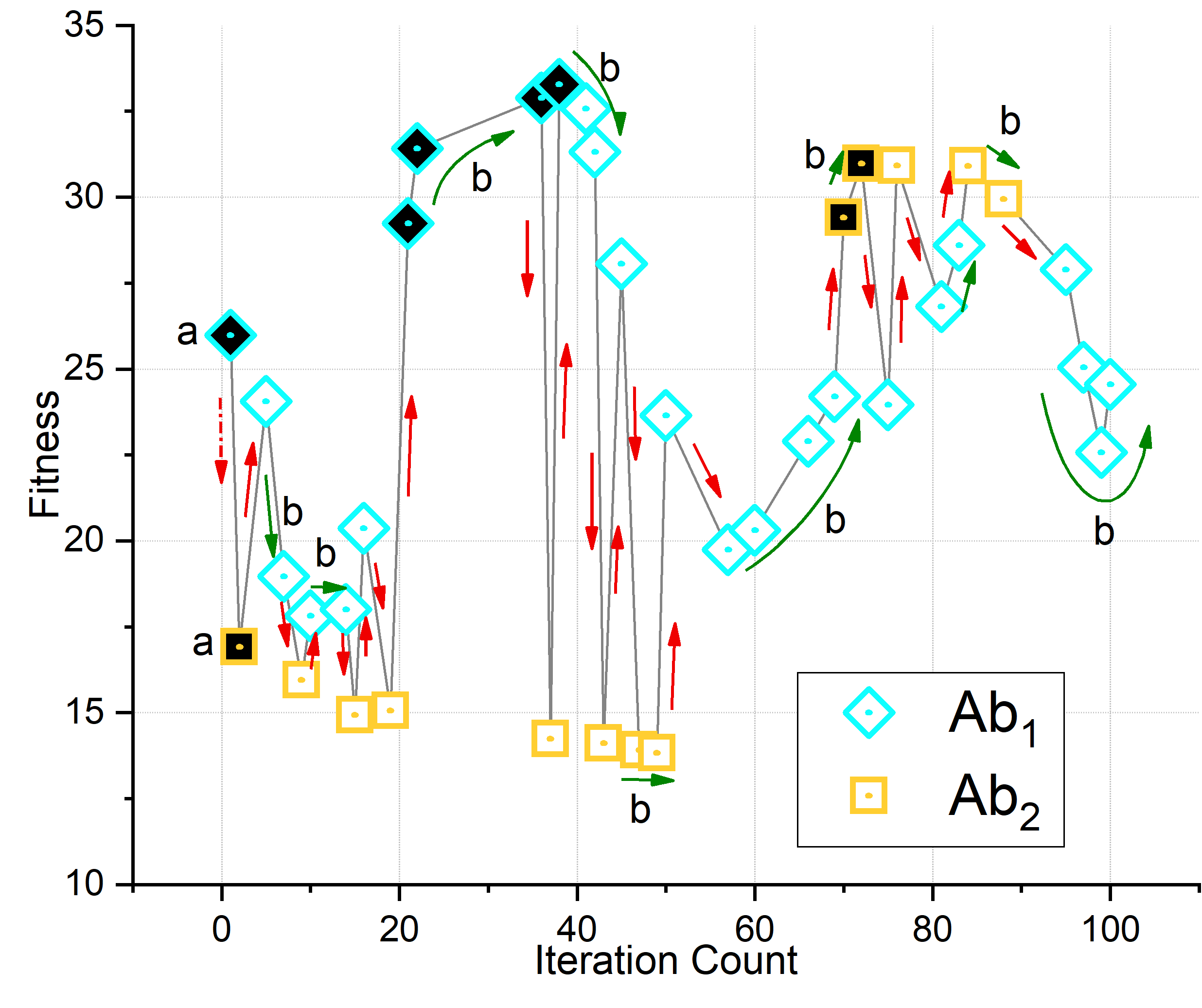} \label{fig:s1}} \
\subfloat[Scenario $S_2$]{\includegraphics[width = 58mm]{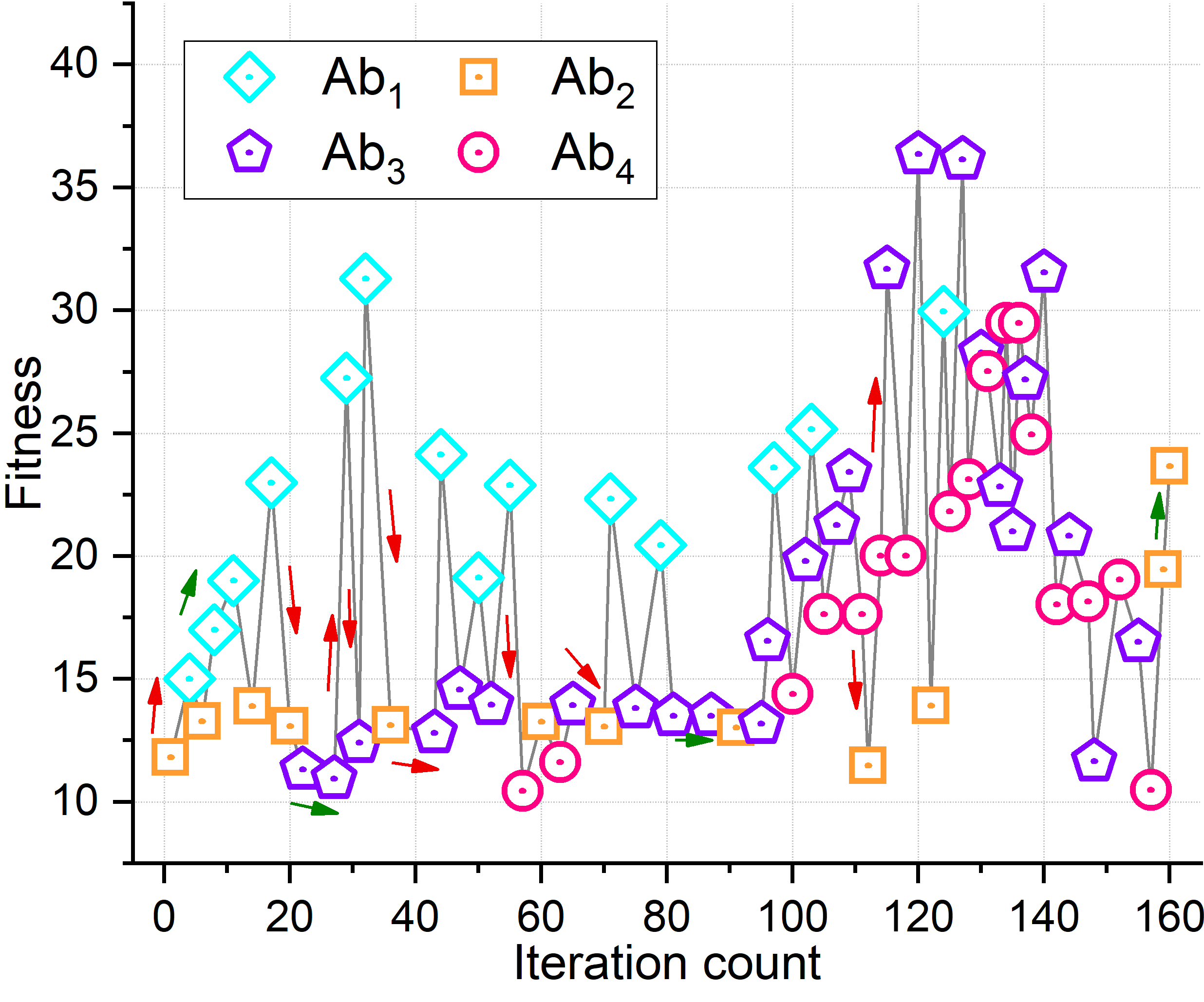} \label{fig:s2}} \
\subfloat[Scenario $S_3$]{\includegraphics[width = 58mm]{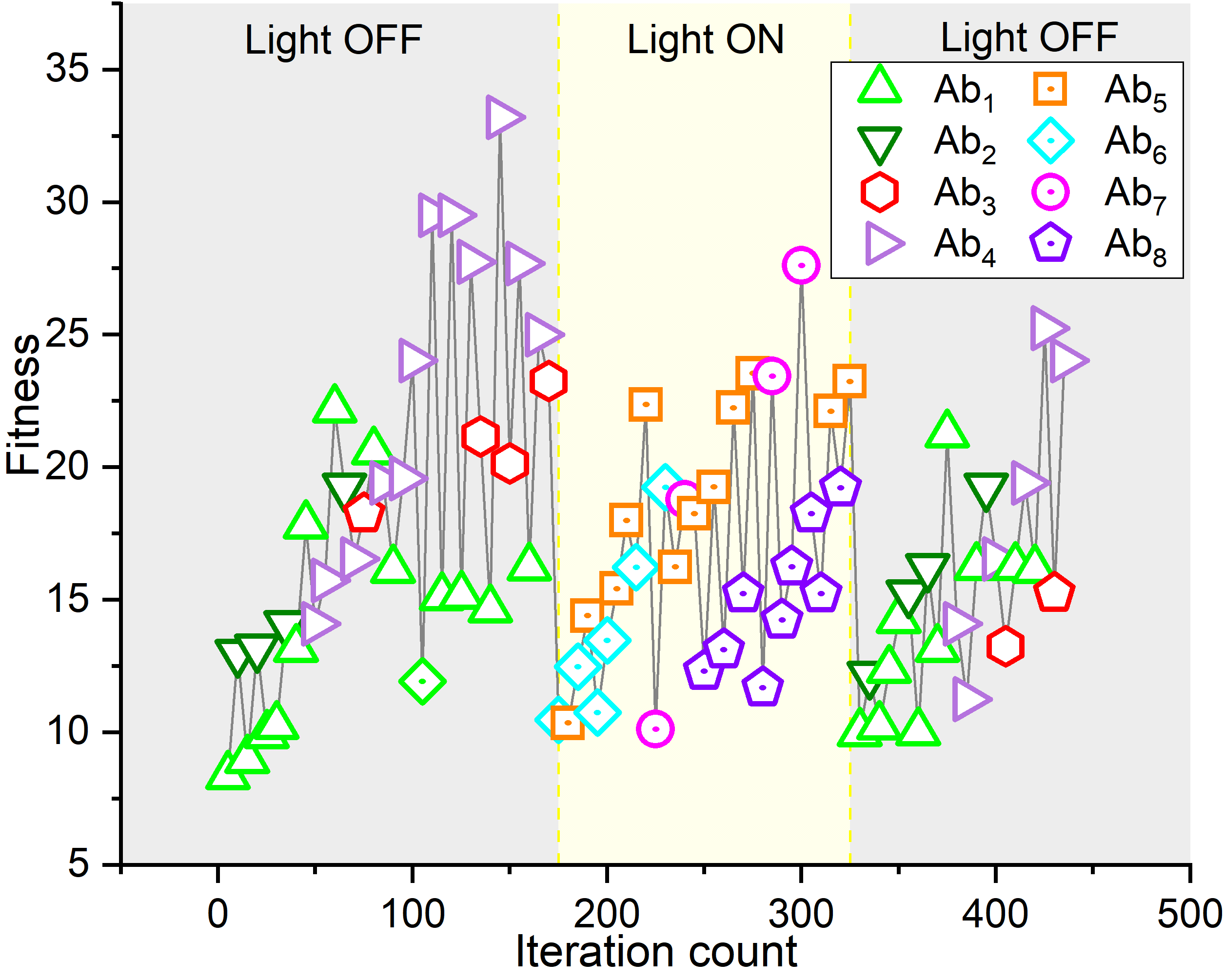} \label{fig:s3}}
    \caption{ Variations in antibody fitnesses for scenarios (a) $S_1$ (b) $S_2$ (c) $S_3$}
    \label{fig:exp}    
\end{figure*}

\section{Results}
In this section, we initially discuss the results on our attempts to evolve a single controller for all the tasks, followed by an analyses of the set of experiments which are performed using the real-robots. Finally, we showcase the results obtained with the IPM scheme running on an emulated network of 100 nodes.\\
\textbf{Evolving a Single Robot Controller}~
A total of ten runs per scenario were performed to evolve a single controller using (1+1)-Online EA running on a single robot. The  controller was allowed to evolve for 200 iterations in scenario $S_1$ while the same was 400 for the other two scenarios. In scenario $S_1$, the controllers which learned to move straight towards the light or avoid the obstacle, easily evolved during the first half of the total of 200 evaluations in all the ten runs of the same experiment. It was only after an average $145^{th}$ evaluation count that controllers which learnt partial phototaxis together with obstacle avoidance emerged. These controllers were of inferior quality in the sense that they followed a curved path while moving towards the light source, instead of the preferred straight movement.

In the scenario $S_2$, a controller which learned all the three subtasks, namely foraging, phototaxis and obstacle avoidance, was not evolved.
From the pool of controllers found during the 400 evaluations, some of the evolved controllers learned only obstacle avoidance while some others learned foraging and phototaxis. No controller learned all the tasks. Scenario $S_3$ had two subtasks which are switched based on the two conditions: 1) Condition $C_1$, when the light source is turned ON, and the robot had to push the pucks towards the light while also avoiding obstacles. 2) Condition $C_2$, wherein the light source is turned OFF and the robot had to learn to avoid obstacles and to repel the pucks encountered. The conditions $C_1$ and $C_2$ were triggered asynchronously. The duration for which each condition lasted was kept such that a minimum of 5 consecutive controller evaluations could take place under each condition. This asynchronous switching made it impractical for a single controller to learn the whole task.
We may thus conclude that it is difficult for a single monolithic controller to learn all the tasks described herein. \\
\renewcommand{\thefootnote}{\faYoutubePlay}
\textbf{Real-Robot Experiments}~We performed 10 trials for each of the scenarios\footnote{For the video: \url{https://goo.gl/8qtJtd}}. The value of $\epsilon$ was empirically found and set to 0.45, 0.4 and 0.25 for the scenarios $S_1$, $S_2$ and $S_3$, respectively. For a given scenario, the desired value of $\epsilon$ can be found by first initializing it to half the normalized range. The value may then be gradually  increased or decreased based on whether the number of subcontrollers generated is sufficient to make the robot(s) learn the scenario. Using the proposed algorithm, the subcontrollers were allowed to evolve for 200 iterations in scenario $S_1$ while the same was 500 for the other two scenarios.\\
\textit{1) Scenario $S_1$}: Fig. \ref{fig:s1} shows a typical evolution-selection curve of the antibodies created, evolved and selected during one of the trials. The X-axis denotes the iteration count and Y-axis indicates the fitness values returned by the objective function defined in equation \ref{eq:t1}. 

For this particular run, the robot was initially placed near the boundary wall of the arena, facing the light source. Following is a description of the events that resulted in the graph shown in Fig. \ref{fig:s1}. The robot initially sampled the $Ag$ from its environment and the first antibody $Ab_1$ was created, depicted by the cyan coloured marker on the graph (Fig. \ref{fig:s1}). The robot then executed the randomly initialized $Ctr$ of $Ab_1$ which caused it to move to a place whose $Ag$, when sampled, was found to be outside the $AR$ of $Ab_1$. The random behaviour executed by the robot was that of rotating around its position. Since initially, the robot was facing an open area, rotation around its axis caused it to face the wall leading to a drastic decline in their US and LS values, justifying the need to generate a fresh $Ab$ to tackle this new and unknown antigenic space. Having sampled this new $Ag$ (vector having small values in the US and LS fields) the robot
created a corresponding fresh antibody $Ab_2$ (orange coloured marker) and executed it. These very first versions of the antibodies $Ab_1$ and $Ab_2$ corresponds to the case (a) in Fig. \ref{fig:shapeSpace}, which shows the creation of new antibodies coincident on their corresponding $Ag$s.  The transitions from $Ab_1$ to $Ab_2$ (and vice-versa) are shown using red coloured arrows on the graph in Fig. \ref{fig:s1}. These transitions from one $Ab$ to another indicate the \textit{on-the-fly} selection property of our proposed algorithm. The green coloured arrows denote the selection of the same $Ab$ after an iteration ends. This indicates that the robot has sensed the next $Ag$ within the AR of the same antibody. 

In both cases (pointed by the arrows), the evolution of an $Ab$ is carried out using an EA where the parent $Ctr$ of $Ab$ is mutated to produce an offspring which is then executed based on a probability. If the offspring perform better than the parent, the offspring $Ctr$ replaces the parent $Ctr$ of the same $Ab$.  Since the robot is in continuous motion, the sampled $Ag$ is always changing. Thus, the same $Ab$ is selected when different $Ag$s appear within its $AR$, as previously highlighted in case (b) of Fig. \ref{fig:shapeSpace}. The best antibodies evolved are indicated with the markers filled with black colour as shown in Fig. \ref{fig:s1}. Iteration 3 onwards, $IRep$ has two antibodies from which the robot could choose one. These include $Ab_1$ which can be triggered for the antigenic space wherein the robot is in an open area and  $Ab_2$ which can be chosen when there is an obstacle in front. This process of evolution and selection continues until the training period ends. The best antibodies evolved can then be used during the testing phase. For the remaining 9 out of 10 runs, the robot is placed in different positions and orientations. It may be noted here that the concept of action-selection should not be confused with those in incremental learning approaches \cite{bongard2008behavior}, as it may seem so from Fig. \ref{fig:s1}. The antibodies herein are selected based on an event without any human interference or for that matter any other entity. \\
\textit{2) Scenario $S_2$}: This scenario is a complex extension of $S_1$ wherein along with phototaxis and obstacle avoidance, the robot also needs to learn to push the puck towards the light source as and when it encounters one. In addition, the robot needs also to learn to avoid an obstacle while pushing this object without mislaying or losing them. Fig. \ref{fig:s2} shows a similar evolution-selection curve (as in Fig. \ref{fig:s1}) wherein the antibodies were evolved and selected during the training period. In this trial, the robot was randomly placed in a position facing the wall of the arena. The repertoire, $IRep$ is set to NULL before initiating the algorithm. Since this scenario is an extension of $S_1$, the antibodies that could tackle the antigenic spaces involving obstacles and phototaxis, emerged to be of similar nature as found during the training in $S_1$. In addition, two new antibodies ($Ab_3$ and $Ab_4$) were created, evolved and selected. $Ab_3$ was selected whenever the robot encountered a puck object with no obstacle in front. $Ab_4$ evolved to tackle the case when the robot encountered an obstacle while pushing the object. The flow of evolution and selection can be seen in Fig. \ref{fig:s2}. The meanings of the red and green arrows are the same as mentioned in the earlier graph.  
It can be seen that while on one side the proposed algorithm evolves new antibodies, it is also capable of selecting a pertinent one amongst these for execution.\\
\textit{3) Scenario $S_3$}: The environment in this scenario changes dynamically and asynchronously with the robot having no control over it. The robot, in turn, needs to change its behaviour based on this change in its environment.  As mentioned in earlier, scenario $S_3$ requires different behaviours, and thus different $Ab$s for each of the light conditions (ON or OFF).
As can be seen from Fig. \ref{fig:s3}, 
the light source was initially in the OFF position from the $1^{st}$ to the $176^{th}$ iteration. After this, it was kept switched ON till the $325^{th}$ iteration and turned OFF again. The $Ab$s specific to the light's OFF state were initially created, evolved and selected during the time the same OFF state was maintained. As soon as the light source was switched ON, the antigenic space changed, leading to the creation of a new set of $Ab$s. These $Ab$s then followed the same evolution-selection journey in order to produce suitable $Ctr$s. As can be seen from the figure, when the light OFF state occurs again, the previously evolved $Ab$s present in the repertoire were triggered and again and evolved further. It may be noted that the $Ab$s in all the runs of each of the experiments for scenarios $S_1$, $S_2$ and $S_3$, were learned from scratch.
\begin{figure}[t]
    \centering
    \includegraphics[scale=.25]
    {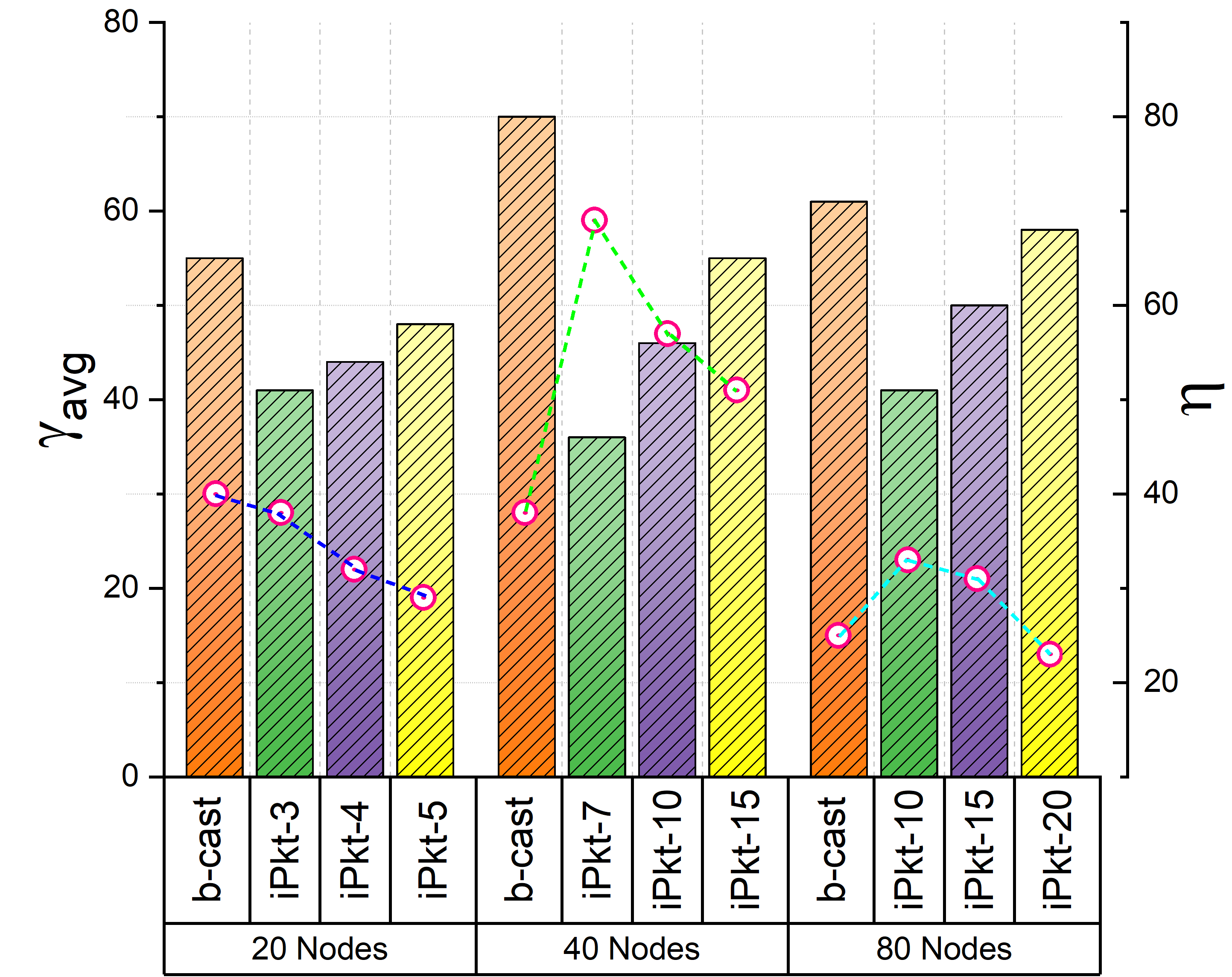}
    \caption{Plots of $\gamma_{avg}$ and $\eta$ for different networks for Broadcast and IPM schemes}
    \label{fig:bcast_ipm}
\end{figure}
A total of 5 independent trials for $S_3$ were also conducted using three real robots which were allowed to share their respective $IRep$s among the peers with similar results. Increasing the number of robots further will aid a parallel search, which in turn can enhance the learning. This experiment also validated that the algorithm can be run across a collective of robots. The advantage of sharing, however, is put down to an extent by the energy consumed in the sharing operations. In the next set of experiments, we propose and implement an alternative way to share the knowledge among the robots in an energy conservative manner. \\
\textbf{Sharing Scheme for Energy Conservation}~Sharing of information within a collective of robots is traditionally realized using broadcasting on the part of individual robots. This method, though simple and effective, consumes a fair amount of energy. Energy conservation is vital in mobile robot scenarios since charging points may not be available or could be far away from its current location. Thus, every attempt to reduce energy consumed needs to be made. We propose a new \textit{Intelligent Packet Migration} (IPM) scheme that is more conservative in energy consumption than the conventional broadcast method.
For experimentation, an emulation environment \textit{Tartarus} \citep{Semwal:2015:TMP:2783449.2783469,Semwal:2016:TMP:2936924.2937224}  was chosen instead of simulation since the former allows for better analyses of real network parameters such as data transfer speed, bandwidth and energy consumed. 

In the IPM scheme, each message packet is a piece of code which has the ability to migrate from one node to another in an autonomous manner. These $Intelligent Packet$s (\textit{iPkt}) can migrate from one node to another in a network, take decisions and also execute the code they carry based on predetermined conditions. An \textit{iPkt} can also carry information in the form of its payload and transport the same to the other nodes during its sojourns across the network. We have used these \textit{iPkt}s to carry the controllers (weights of the ANN) evolved by a robotic node to facilitate sharing with its peers in the network. To restrict extraneous movement of such a packet, we programmed the same to move in a \textit{conscientious} manner wherein the \textit{iPkt} maintains a list of already-visited robot nodes. Every time an \textit{iPkt} visits a node, it appends the node identifier (node address) in the list. In the conscientious movement,  an \textit{iPkt} migrates to that neighbouring node which is either not visited or has been least visited. Unlike broadcasting, the conscientious method provides for a controlled movement and thus greatly aids in the reduction of unnecessary packet migrations. In addition, an \textit{iPkt} can also be programmed to move to another robotic node under certain conditions such as when a controller with a desired fitness is found. This could further contribute to reducing the energy spent on communication. It may be noted that to know the neighbours, an \textit{iPkt} at a node needs to broadcasts a small \textit{hello} message packet to create a routing table. Since the \textit{hello} broadcast is done only by those nodes that currently have a packet within and when it wants to migrate to a neighbouring node, the overall communication overheads turn out to be less than that in case of broadcasting.
For comparing the efficacy of the sharing schemes, we have used (and implemented) a standard EA such as the one proposed in \cite{heinerman2016line} to solve the soft task of learning an OR gate function. An ANN with 2 input nodes, 3 hidden nodes and 1 output node was used. For broadcasting, we used the social learning method prescribed in \cite{heinerman2016line} while in the case of IPM, we implemented the encapsulated version (individual learning) of the same method as in \cite{heinerman2016line} and used the \textit{iPkt}s to share the evolving controllers among the emulated robot nodes forming the network.

In Fig. \ref{fig:bcast_ipm}, the left Y-axis denotes the inter-robotic node communications (sending and receiving) per robot node averaged ($\gamma_{avg}$) over 10 runs of the experiments. The right Y-axis is based on a metric called the \textit{Convergence count} ($\eta$), which is the number of iterations spent from the point when the first best solution was found by a node till 90\% of the robot population converges to the same solution due to sharing. In the X-axis, \textit{b-cast} denotes the broadcast method while the terms $iPkt$-$x$ correspond to the cases when $x$ \textit{iPkt}s were used. The value $x$, i.e. the number of $iPkts$ is chosen proportionately to the total number of robotic nodes in the network. The emulation experiments were carried out on 20, 40 and 80-node dynamic networks. Dynamism was introduced by varying the number of neighbours per node from 0 (isolated node) to a maximum of 5. The neighbours of a node were changed randomly over time making it the equivalent of a dynamic network.

As can be seen from Fig. \ref{fig:bcast_ipm}, with an increase in the number of nodes, the average number of communications made per node using the broadcast method remains higher than that reported using the IPM scheme. 
The graph also shows that as the number of \textit{iPkt}s is increased, the sharing too is enhanced (lower convergence counts). This can be observed for each of 20, 40 and 80-node scenarios where the convergence counts drops with the increase in \textit{iPkt}s. It may be noted that, though one may infer that an increase in \textit{iPkt}s would accelerate the sharing process, it comes at the cost of higher communication overheads per node. Overall, one may thus conclude that if saving power is prime, then the IPM method seems to have an edge over the conventional broadcast mechanism.

\section{Discussions and Conclusions}
This paper describes an immunology-inspired distributed and embodied action-evolution cum selection algorithm for learning of specific subcontrollers in an online manner. Robots in a collective evolve individual subcontrollers for coping up with the environment just the way the BIS creates and tunes antibodies to quell antigenic attacks on-the-fly. Subcontrollers are shared amongst the robots to facilitate other robots to get better solutions and cope up with the environment. Based on the value of $\epsilon$, the environment (antigenic) space sensed by the robot can be partitioned into different \textit{Active Regions} (AR). A separate subcontroller is then evolved for each AR. Thus, $\epsilon$ governs the number of subcontrollers generated. More experiments will need to be carried out to comprehend the manner of selecting an appropriate value for $\epsilon$ for a given environment.     
Since power can be a major source of concern in mobile robots, the use of \textit{iPkt}s for sharing allows for a fair saving in the same. Further investigations into adjusting the number of \textit{iPkt}s and their subsequent routing will provide better insights into making the sharing mechanism more energy efficient.

\bibliographystyle{ACM-Reference-Format}


%\bibliography{biblio} 

\end{document}